\def\year{2020}\relax
\documentclass[letterpaper]{article} % DO NOT CHANGE THIS
\usepackage{aaai20}  % DO NOT CHANGE THIS
\usepackage{times}  % DO NOT CHANGE THIS
\usepackage{helvet} % DO NOT CHANGE THIS
\usepackage{courier}  % DO NOT CHANGE THIS
\usepackage[hyphens]{url}  % DO NOT CHANGE THIS
\usepackage{graphicx} % DO NOT CHANGE THIS
\usepackage{graphicx}  % DO NOT CHANGE THIS
\title{Examining Racial Bias in an Online Abuse Corpus with Structural Topic Modeling}
\author{Thomas Davidson \textnormal{and} Debasmita Bhattacharya \\
Cornell University \\
trd54@cornell.edu, db758@cornell.edu}
\begin{document}
 
\maketitle

\frenchspacing % Uncomment to make text smaller

\begin{abstract}
We use structural topic modeling to examine racial bias in data collected to train models to detect hate speech and abusive language in social media posts. We augment the abusive language dataset by adding an additional feature indicating the predicted probability of the tweet being written in African-American English. We then use structural topic modeling to examine the content of the tweets and how the prevalence of different topics is related to both abusiveness annotation and dialect prediction. We find that certain topics are disproportionately racialized \textit{and} considered abusive. We discuss how topic modeling may be a useful approach for identifying bias in annotated data.
\end{abstract}

\section{Background}
There has recently been a surge in attention to the problem of online abuse and hate speech. There is a growing literature on the characteristics of this speech and how to detect it, including several annotated datasets \cite{waseem_hateful_2016,golbeck_large_2017,davidson_automated_2017,founta_large_2018}.\footnote{See \url{http://hatespeechdata.com/} for an up-to-date list of available datasets.} However, recent work finds that these datasets contain bias against African-Americans, insofar as tweets written in dialect used by African-Americans are often considered more hateful or offensive \cite{davidson_racial_2019,sap_risk_2019}. Given the ubiquity of bias in both supervised and unsupervised machine-learning models, it is likely that other biases are also present \cite{bolukbasi_man_2016,blodgett_racial_2017,caliskan_semantics_2017,tatman_gender_2017,garg_word_2018,kiritchenko_examining_2018}.

In this paper we further investigate this issue by looking more closely at the content of a large set of tweets annotated for abusive language and hate speech \cite{founta_large_2018}. We use a language model \cite{blodgett_demographic_2016} to predict the probability that tweets in the dataset are written in African-American English (AAE) \cite{sap_risk_2019}. We then use these predictions and the abusiveness annotations as covariates in a structural topic model \cite{roberts_structural_2014}. This enables us both to better understand the latent topics in the corpus and to see what types of language are associated with both annotations for abusiveness and AAE. The vast majority of topics exhibit no association between topic prevalence and AAE usage. We identify two topics that appear to be particularly prevalent in abusive tweets and find that one of them is strongly associated with AAE usage. This demonstrates how topic modeling can serve as a useful tool to further uncover biases in annotated text data.

\section{Data}
We use the Task 2 dataset for the ICWSM Data Challenge focused on abusive behavior identification \cite{founta_large_2018}. The dataset consists of 99,996 tweets annotated by human raters into four categories: hateful, abusive, spam, and normal. To approximate a realistic scenario where abusive language is relatively rare, the dataset contains a large random sample of tweets along with a subset of tweets with negative sentiment and one or more offensive words. We drop tweets annotated as spam, focusing on the main three categories. While we recognize that the distinction between hate speech and other kinds of abusive language is important \cite{davidson_automated_2017}, we collapse the two categories into a single category.

% TODO: Add an online Appendix to the Github repo containing: (a) Diagnostic plot, (b) list of topic names and examples, (c) our topic classifications. for the Arxiv version I just link to Github
\section{Methodology\footnote{Replication materials are available on Github: \url{https://github.com/db758/icwsm_data_challenge}}}
We start by predicting the probability that each tweet in the dataset is written in African-American English. Following \cite{sap_risk_2019}, we use a pretrained language model \cite{blodgett_demographic_2016}. This returns a vector of probabilities that a tweet is written in Standard American English, AAE, or in Asian or Hispanic language models. Here we focus on the AAE category\footnote{The Asian and Hispanic models are also not as reliable and the authors of the original paper recommend that they not be used \cite{blodgett_demographic_2016}.}. Some tweets were dropped from the analysis after this step because they did not contain any tokens that could be used by the classifier (e.g. the content was entirely mentions or URLs). 

Next, we train a structural topic model (STM) to identify latent topics in the corpus. As with standard Latent Dirichlet Allocation, this procedure estimates a probabilistic model that describes the latent topics in a corpus of text. Each document, in this case each tweet, is comprised of a distribution over these latent topics. Unlike the LDA model, the STM allows the addition of covariates into the estimation process, which alter the prior distributions used by the model \cite{roberts_structural_2014,lucas_computer-assisted_2015}. In this case, we expect that certain topics might be more prevalent depending upon either the abusiveness or the AAE content of the tweet. We estimated the STM using the \texttt{stm} package in \textsf{R} \cite{roberts_stm:_2019}, specifying a multiplicative relationship between the abusiveness category and the probability of being in AAE as a topic prevalence covariate. 

There is a large number of duplicate documents in the corpus, which can impact the performance of topic models \cite{schofield_quantifying_2017}. We therefore dropped 10,679 duplicate tweets, retaining the first instance of each. We replaced all \@mentions and  URLs with placeholder tokens. Before training the topic model, we further pre-processed the tweets using \texttt{textProcessor} function in \texttt{stm}. This removes punctuation, numbers, and stopwords and sets all text to lowercase. We drop any terms that occur in fewer than 5 tweets\footnote{This removes 57,392 of 65,961 terms used in the corpus. The remaining 8569 terms account for \~90\% of the tokens in the original corpus.} and five documents which were composed solely of such terms. The final analytic sample consists of 75,023 tweets, 32\% of which are labelled as abusive or hateful.

% TODO: Move the image to Github.
A challenge when estimating any topic model is to identify parameter $K$, the appropriate number of topics. Since the corpus is relatively large and consists of a heterogeneous sample of social media content we expect that there could be many different topics present. We experimented with different values of $K$, ranging from 10 to 60. We evaluated the results using quantitative diagnostics\footnote{The diagnostic plot produced by the \texttt{searchK} function can be viewed here: \url{https://imgur.com/a/UdK3pkR}. The models are scored on four different metrics. No one model scored best on all four metrics, but the model with 30 topics appeared to be the best trade-off between them. For discussion of these metrics refer to \cite{roberts_stm:_2019}.} and by qualitatively inspecting the results. Based on this analysis we decided use a model where $K = 30$. After we estimated the model with 30 topics, each of the authors independently annotated each of the topics using the words most strongly associated with each topic and the ten tweets containing the highest proportion of each topic. We compared and discussed our annotations to develop a common interpretation of the topics. Below we focus on a subset of the topics that we considered most relevant to our research aims. 

We used the \texttt{estimateEffect} function in \texttt{stm} to run a linear regression to estimate the proportion of each topic present in a tweet as a function of the annotation, the predicted probability the tweet is written in AAE, and the interaction between these two variables. This allows us to examine how the prevalence of topics varies depending upon whether or not a tweet is abusive or is likely to be written in AAE.

\section{Results}
The corpus contains a wide array of different topics, which is to be expected given that many of the tweets were randomly sampled \cite{founta_large_2018}. Of the 30 topics, we were able to agree upon names for 23: relationships and emotions; school and college; careers and recruitment; cursing; weather; blogs and journals; interior design; sports; war and conflict; art and design; music and television; religion and spirituality; American politics; Twitter follower update bots; cursing and porn; marketing and strategy; celebrity news; thanks and compliments; fashion and online shopping; food; click-bait news; and Twitter mention spam. The remaining 7 topics were less intelligible, as is typical in such analyses \cite{karell_rhetorics_2019}.\footnote{A full list of topics and example tweets is available on Github (link above).}
% TODO: Move these topics to Github.

Here we focus on four topics in particular, the two we considered to be the most abusive and two more normal topics. Topic 4, cursing, is characterized by words including ``*ss'', ``b*tch'', and ``n*gga'' - words shown to be associated with AAE tweets that are annotated as hateful or abusive \cite{davidson_racial_2019,sap_risk_2019}. Tweets with a high proportion of this topic include \textit{``That's why you neva beef ova h*es bra they be f*cking wit everybody''}, \textit{``i hate ole dusty *ss mississippi but jus for these moments i gotta be there''}, and \textit{``These bad b*tches always b wit sum lame bum *ss n*ggas''}.\footnote{Mentions, URLs, and emojis have been removed from examples and curse words obscured.} While some of the words used are potentially offensive, it is not clear that any of these tweets are intended to be abusive or hateful \cite{davidson_automated_2017}, although we did find one example of a racist attempt to imitate AAE, which included the string \textit{``N*gga fried chicken n*gga watermelon n*gga''}. Topic 20, cursing and porn, contains other curse words like ``f*ck'', ``*sshole'', and ``c*nt''. The tweets with the highest proportion of this topic appear to be a mixture of angry expressions (e.g. \textit{``I hate hate hate hate hate hate HATE animal cruelty. That sh*t p*sses me the f*ck off bro. ''}), personal attacks (e.g. \textit{``I f*cking hate you i literally wanna kill you so f*cking bad you fat stupid b*stard''}), and references to pornography (e.g. \textit{``Chubby Busty Redhead Sl*t Sucking And F*cking''}). Turning to the more normal topics, we chose to examine topic 12, which contains tweets related to the Syrian civil war and a range of other conflicts around the world, and topic 24, which contains thanks and compliments and seems to be the most positive in terms of sentiment of all the topics. We expect that topic 12 may be associated with abuse given the contentious nature of the issues discussed in the example tweets and that Topic 24 should be one of the least likely topics to be associated with abuse. These more normal topics serve as a baseline against which we can compare the more abusive topics.

\begin{figure*}[ht]
\center
\includegraphics[width=1\textwidth]{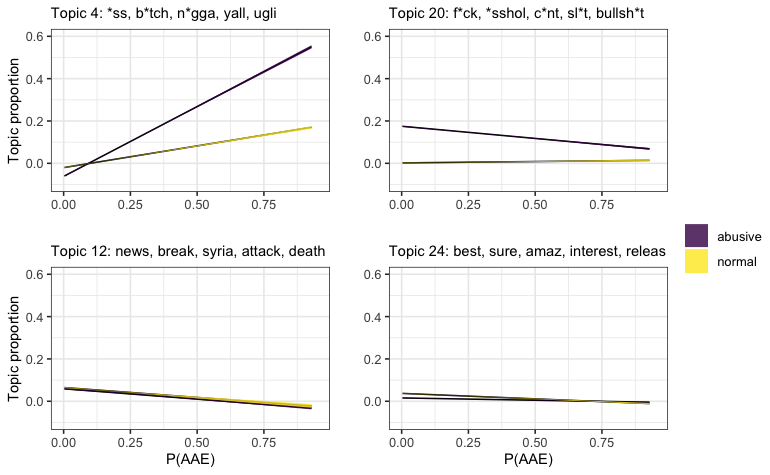}
\caption{Expected topic proportion by probability tweet is written in African-American English for abusive and normal tweets.}\label{fig:f1}
\end{figure*}

Figure~\ref{fig:f1} shows the results of the regression analyses for these four topics. Each panel shows the relationship between the estimated topic proportion and the probability a tweet is written in AAE. The purple line shows the relationship for all tweets annotated as abusive or hateful, the yellow line for those considered normal. Lines include 95\% confidence intervals. Starting with the top left panel, there is a positive relationship between the probability a tweet is written in AAE and the proportion of the tweet using topic 4. This relationship is particularly strong in tweets annotated as abusive. Tweets considered to have a high probability of being written in AAE are expected to contain a relatively large proportion of topic 4. For example, if P(AAE) is around 80\% then we can expect more than 50\% of a tweet to be from topic 4. Thus, there is a strong association between a tweet being written in AAE and containing a high proportion of topic 4, and this association is particularly strong for abusive tweets. In the top right of the figure we see a different pattern for topic 20. The slope of the normal line is nearly horizontal at zero, indicating that tweets considered to be normal contain a negligible amount of text drawn from topic 20. For the abusive tweets we see a negative slope: the higher P(AAE), the lower the expected proportion of the topic. In sum, these two figures show how the prevalence of certain abusive topics is associated with both AAE and abusiveness.

Turning to the bottom two panels, on the left, the proportion of topic 12 declines towards zero as we increase the probability that a tweet is written in AAE. The relationship is nearly identical regardless of whether a tweet is annotated as abusive or normal, suggesting that this category may not be particularly useful as an indicator of abusiveness. On the right, we see a similar negative trend for topic 24, where both lines are relatively flat. This shows that there is a very weak association between P(AAE) and the proportion of a tweet containing this topic. We examined these plots for all topics in the model. The vast majority tended to resemble these bottom two panels, with only weak associations between topic proportion and P(AAE) and little difference between abusive and normal tweets.

\section{Discussion and Conclusions}
% Note how topic modeling can be used as a granular way to understand large corpora of social media data.
We have explored how structural topic modeling can be used to understand the relationship between content and human annotations in a large corpus of annotated social media data. By incorporating covariates into the topic model, we can see how certain topics tend to be disproportionately associated with content annotated as abusive or hateful. In this case, only two topics appear to be strongly associated with these negative categories. One of the topics, topic 4, is strongly positively associated with AAE, while the other, topic 20, has a weaker negative association. Thus, this inductive approach allows us to better understand potential sources of bias identified in this corpus in previous work \cite{davidson_racial_2019,sap_risk_2019}. Moreover, the topic model specified with an interaction between dialect and abusiveness allows us to disaggregate potentially abusive tweets into different clusters. This approach may thus be a useful way to identify false positives - content that is flagged as abusive but composed on topics not associated with abuse - and false negatives - content not flagged but that contains high proportions of topics associated with abuse. Overall, this work demonstrates how unsupervised learning methods can be used to uncover hidden structures in annotated corpora of text data, and how these latent factors are associated with both the annotations we use to train predictive models and the social categories that can result in algorithmic bias and discrimination.

\subsection{Limitations}
First, due to space constraints we grouped together abusive and hateful tweets and so do not assess how these results vary if we disaggregate these categories. Second, while the STM clearly retrieves some signal from the dataset, as most of the topics were considered to be coherent by two independent annotators, it is not ideal to use such short documents. While this work serves as a proof-of-concept, further work should consider alternative approaches better optimized to the medium. Third, we used a pre-trained model to predict whether tweets are written in AAE. It is possible that this model relies upon similar cues as the human annotators, amplifying the association between AAE and abusiveness. Fourth, it is beyond the scope of this paper to examine how this technique can be used to mitigate these biases. Finally, we have only looked at one particular type of racial bias. It is likely that other types of bias are also present such datasets. We hope these limitations can be addressed in future work.

\bibliography{aaai20}
\bibliographystyle{aaai}
\end{document}